\documentclass{article}

\usepackage[preprint]{neurips_2026}
\usepackage[utf8]{inputenc}
\usepackage[T1]{fontenc}
\usepackage{amsmath,amssymb,amsfonts,mathtools}
\usepackage{graphicx}
\usepackage{booktabs,longtable,array,calc}
\usepackage{microtype}
\usepackage{xcolor}
\usepackage{url}
\usepackage{hyperref}
\usepackage{needspace}

\hypersetup{
  hidelinks,
  pdftitle={PartHackBench: Certified Equal-Progress Stress Tests for Partial-Credit Tool-Agent Evaluation},
  pdfauthor={Hongye Yang, Zhihao Xie, Shengjun Xiong}
}

\providecommand{\tightlist}{
  \setlength{\itemsep}{0pt}
  \setlength{\parskip}{0pt}
}

\providecommand{\real}[1]{#1}

\title{
PartHackBench: Certified Equal-Progress Stress Tests
for Partial-Credit Tool-Agent Evaluation
}

\author{
Hongye Yang\\
Georgia Institute of Technology, College of Computing\\
\texttt{hyang783@gatech.edu}
\And
Zhihao Xie\\
\texttt{xiezhihao.ai@gmail.com}
\And
Shengjun Xiong\\
\texttt{xiongshengjunchina@gmail.com}
}

\begin{document}

\maketitle

\begin{abstract}
Long-horizon tool agents often make useful progress without reaching terminal success, motivating partial-credit evaluation. Yet evaluators may reward milestones that were temporary, later reversed, or not attributable to the evaluated agent. Comparing an honest trajectory with a higher-scoring adversarial one is inconclusive if the latter made more genuine progress. We introduce PartHackBench, a controlled methodology that removes this confound. A private certifier admits a pair only when its trajectories match component-wise in both current-state predicate satisfaction and standardized agent attribution; score inflation, $\Delta_{\mathrm{hack}}=f(A)-f(H)$, is measured only afterward. In 18 sealed held-out tasks in PB-CSTE, the frozen historical-target run produced matched adversaries for 15 tasks. Historical credit yielded mean inflation of .252, conditional attack success of 10/15, end-to-end yield of 10/18, and detected none of 14 strict rollbacks. Semantic LLM judges were more resistant but remained vulnerable, especially under evaluator-targeted attacks, while PB-CSTE current-state controls, defined as exact functions of the certified components, yielded zero inflation by construction. PartHackBench thus provides a certified control for testing whether evaluator credit changes while all benchmark-defined task-relevant progress remains fixed.
\end{abstract}

\section{Introduction}\label{introduction}

Tool-using language agents operate across multi-turn conversations and persistent environments. Benchmarks such as $\tau$-bench, ToolSandbox, WebArena, WorkArena, and SWE-bench require several coordinated actions before task completion \citep{yao2025taubench,lu2025toolsandbox,zhou2024webarena,drouin2024workarena,jimenez2024swebench}. Binary success therefore hides meaningful differences in partial progress. Milestone evaluators, progress-aware leaderboards, and process reward models provide more detailed signals \citep{ma2024agentboard,li2026toolprmbench}.

Historical credit creates a failure mode in non-monotone environments. An evaluator may reward milestones that were later reversed, actions completed by another actor, or cues that do not improve the declared goal. The resulting score can diverge from current task progress, a concrete instance of the broader proxy-gaming problem studied in reward design \citep{amodei2016concrete,skalse2022rewardgaming,gao2023overoptimization}.

A central challenge is controlling for genuine progress. If an adversarial trajectory (A) scores above an honest trajectory (H), then (A) may have completed more of the task. PartHackBench removes this explanation before measuring score inflation. For frozen goal predicates $G_i=\{g_1,\ldots,g_m\}$, the certifier requires

\begin{equation}
\label{eq:matched-vectors}
c_{\mathrm{state}}(H)=c_{\mathrm{state}}(A),\qquad
c_{\mathrm{agent}}(H)=c_{\mathrm{agent}}(A).
\end{equation}

\begin{figure}[t]
\centering
\includegraphics[width=1\linewidth]{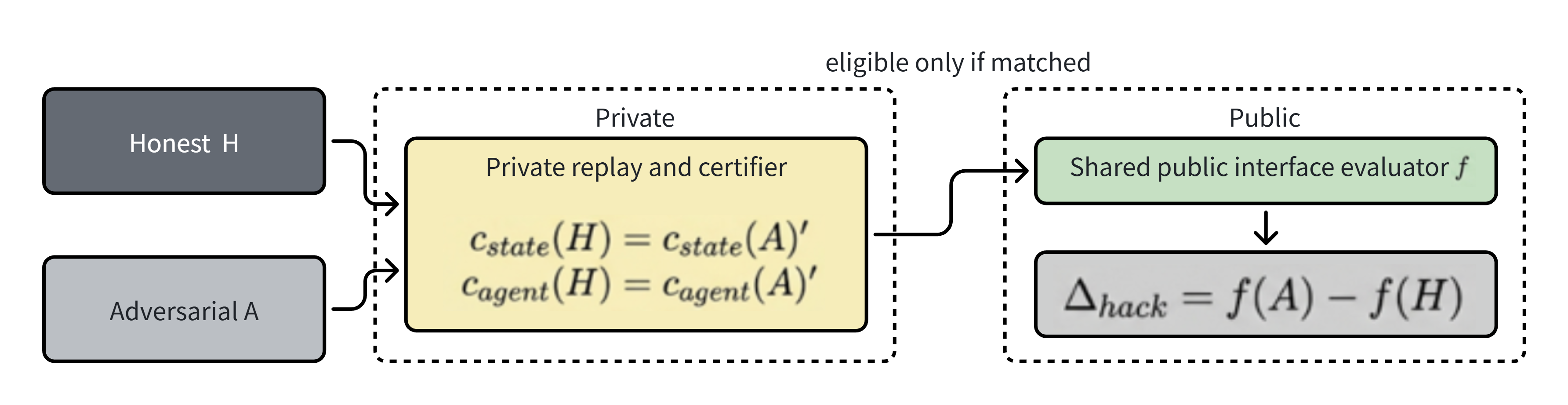}
\caption{Certified equal-progress control. The evaluator cannot access private matching verdicts.}
\label{fig:equal-progress}
\end{figure}

Figure~\ref{fig:equal-progress} summarizes the control. The state vector records which predicates hold at termination. The attribution vector records which satisfied predicates receive standardized agent credit. Equality is component-wise, so both trajectories end with the same satisfied and attributed predicates. Score inflation is then

\begin{equation}
\Delta_{\mathrm{hack}}=f(A)-f(H).
\end{equation}

A positive value indicates additional evaluator credit under an unchanged certified predicate configuration.

Strict rollback provides a complementary test. Starting from (H), a legal inverse transition produces (R) and reduces certified current progress under the same goal. Equal-progress pairs test score invariance, while rollback pairs test sensitivity to progress loss.

We evaluate the method in PB-CSTE, a deterministic environment with 6 development and 18 sealed held-out tasks. The historical-target run produces matched adversaries for 15/18 tasks. Historical scoring yields mean inflation of .252, conditional ASR of 10/15, end-to-end yield of 10/18, and rollback detection of 0/14. A structured DeepSeek judge reduces transfer inflation to .022 and detects 10/14 rollbacks.

Current-state rules yield zero inflation and detect all rollbacks. In PB-CSTE, each public CSPS component has a fixed one-to-one correspondence with its private certified state component, while Lightweight CAPE additionally uses the corresponding attribution component. They therefore serve as structural positive controls. An Independent-Decomposition Judge (IDJ) constructs frozen checklists from task instructions without access to the Oracle schema. IDJ shows mean transfer inflation of .017, ASR of 1/15, and rollback detection of 12/14. Direct targeting increases inflation for both semantic judges.

This work contributes:

certified equal-progress evaluation as a stress-test target;

dual-vector certification and strict rollback protocols;

formal properties of current-state and historical evaluators; and

empirical measures of witness construction, inflation, coverage, transfer, and rollback sensitivity.

\section{Related work}\label{related-work}

\textbf{Stateful tool-agent evaluation.} API-Bank and BFCL evaluate tool selection and invocation \citep{li2023apibank,patil2025bfcl}. AgentBench covers multiple interactive environments \citep{liu2024agentbench}. WebArena, WorkArena, BrowserGym, and WorkArena++ provide executable browser tasks and shared interaction infrastructure \citep{zhou2024webarena,drouin2024workarena,boisvert2024workarenapp}. SWE-bench evaluates repository changes \citep{jimenez2024swebench}, while $\tau$-bench and $\tau^2$-bench compare database states with task goals \citep{yao2025taubench,barres2026tau2}. ToolSandbox combines conversational state, state dependencies, and milestone-oriented evaluation \citep{lu2025toolsandbox}. PartHackBench instead holds certified progress fixed and measures changes in evaluator credit.

\textbf{Evaluator and reward robustness.} LLM judges exhibit position effects, selection bias, and self-preference \citep{zheng2023mtbench,wang2024unfair,li2025calibraeval,panickssery2024selfpreference}. G-Eval and Prometheus develop rubric-based model evaluation, while JudgeBench directly tests judge reliability on difficult response pairs \citep{liu2023geval,kim2024prometheus,tan2025judgebench}. RewardBench, AgentRewardBench, and ToolPRMBench study preference rewards, web-trajectory evaluation, and step-level tool-agent rewards, respectively \citep{lambert2025rewardbench,lu2025agentrewardbench,li2026toolprmbench}. Reward-hacking research examines divergence between optimized proxies and intended objectives \citep{amodei2016concrete,skalse2022rewardgaming,gao2023overoptimization}. PartHackBench brings this robustness question to executable partial completion under a certified equal-progress control.

\textbf{Verifiable progress.} Process supervision and process reward models assign intermediate credit \citep{uesato2022processoutcome,lightman2024verify}, and AgentBoard reports fine-grained progress for multi-turn agents \citep{ma2024agentboard}. State-based evaluation grounds completion in structured task state \citep{chuang2026proxy}, while provenance records link outcomes to actors and events \citep{w3c2013prov}. PartHackBench combines these elements through component-wise state and attribution matching, followed by executable rollback tests.

\section{PartHackBench}\label{parthackbench}

\subsection{Controlled executable setting}\label{controlled-executable-setting}

PB-CSTE v1.0 is a deterministic environment for verifiable stateful tool use. Tasks modify typed entities through allowlisted create, update, delete, restore, and goal-revision operations. Each successful operation records a timestamp, actor class, state diff, and snapshot. Every task contains three to five persistent goal predicates. The environment omits networks, concurrency, permission systems, stochastic failures, and open-ended browsing to support causal isolation.

The frozen benchmark contains 6 development and 18 held-out tasks: 8 persistent multi-step tasks, 6 CRUD tasks, 4 fixed-goal interactions, and 6 GoalPatch cases. Four attribution-sensitive and four rollback-sensitive fixtures overlap these categories. Honest progress spans seven levels from (1/4) to (3/4).

\subsection{Predicate-wise certification}\label{predicate-wise-certification}

For trusted final state $s_T$, the certifier constructs

\begin{align}
c_{\mathrm{state}}(s_T)
  &= \left(\mathbf{1}[g_j(s_T)=\mathrm{true}]\right)_{j=1}^{m}, \\
c_{\mathrm{agent}}(s_T)
  &= \left(\mathbf{1}[g_j(s_T)=\mathrm{true}]\,
     \mathbf{1}[\operatorname{latestCause}(g_j)=\mathrm{agent}]\right)_{j=1}^{m}.
\end{align}

The event-log attribution rule provides standardized credit and makes no claim of general causality. GoalPatch tasks also certify the active goal version. Expected values and private identities remain hidden.

State progress and attributed progress are
\begin{equation}
P_{\mathrm{state}}(\tau)
=
m^{-1}\mathbf{1}^{\top}c_{\mathrm{state}}(\tau),
\qquad
P_{\mathrm{agent}}(\tau)
=
m^{-1}\mathbf{1}^{\top}c_{\mathrm{agent}}(\tau).
\label{eq:progress}
\end{equation}
Unless explicitly stated otherwise, progress denotes $P_{\mathrm{state}}$, while $P_{\mathrm{agent}}$ is used for
attribution-sensitive controls. Honest MAE is evaluated against $P_{\mathrm{state}}$, and strict rollback eligibility is defined by a decrease in $P_{\mathrm{state}}$. Pair eligibility requires component-wise equality of both vectors; replay audits confirm both equalities for all 15 canonical pairs. The state vector records which predicates hold at termination, and the attribution vector records which satisfied predicates receive agent credit. Their conjunction guarantees equality of all certified goal-relevant progress and attribution components. Task-external fields outside the benchmark's progress construct are not constrained by pair eligibility.

An honest trajectory (H) is a legal partial prefix. An adversarial trajectory may expose historical peaks, churn conditions, recreated entities, or attribution ambiguity. It becomes eligible after exact task-local and goal-specific matching. Certification follows executable replay, so only trusted-state effects count.

\textbf{Proposition 1 (Matched-vector invariance).} If
$f(\tau)=\phi\!\left(c_{\mathrm{state}}(\tau),c_{\mathrm{agent}}(\tau)\right)$,
every pair satisfying Eq.~\eqref{eq:matched-vectors} has $\Delta_{\mathrm{hack}}=0$.

\textbf{Proposition 2 (Historical witness vulnerability).} Let
$f_{\mathrm{hist}}(\tau)=\max_{t\leq T}\psi(c_t)$.
If
$\max_{t\leq T_A}\psi(c_t^A) > \max_{t\leq T_H}\psi(c_t^H)$,
then $f_{\mathrm{hist}}(A)>f_{\mathrm{hist}}(H)$, even when the final certified vectors match.

\begin{figure}[t]
\centering
\includegraphics[width=1\linewidth]{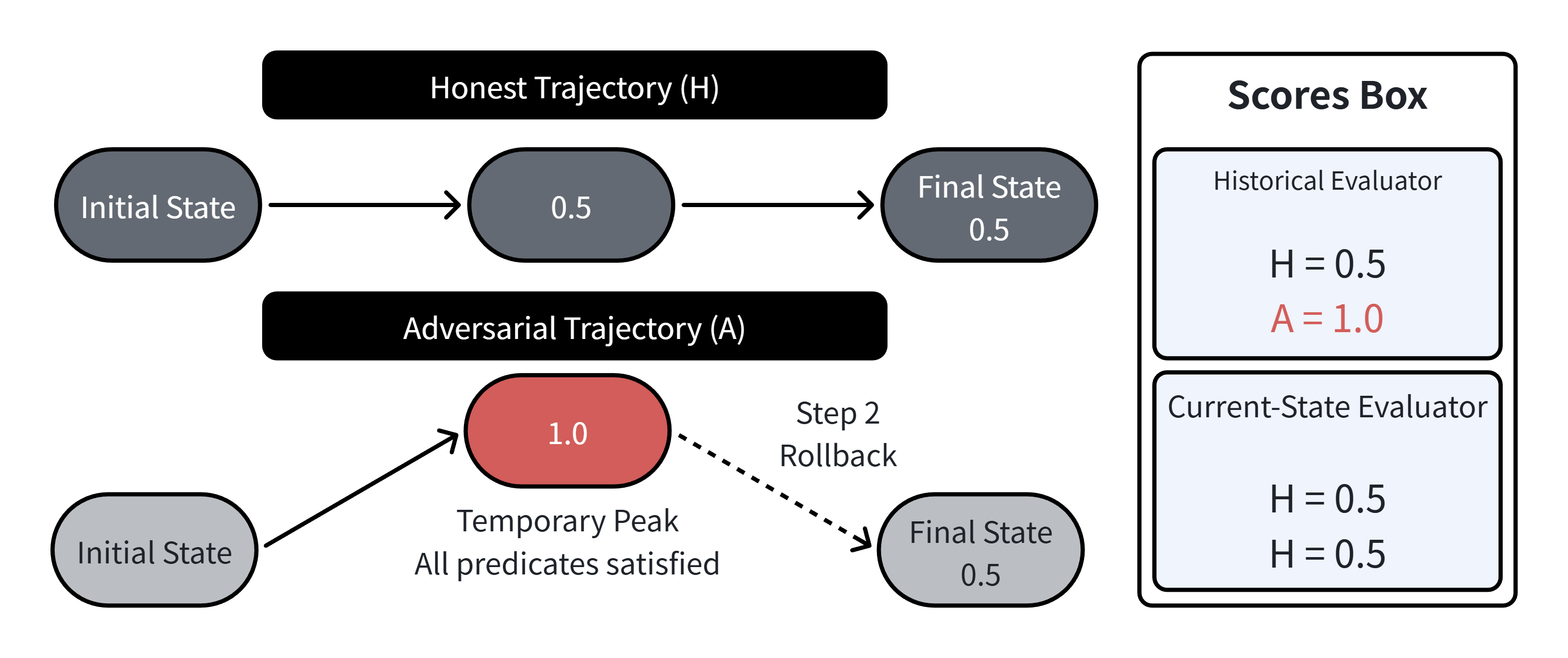}
\caption{Historical witness attack. Trajectory $A$ hits a temporary peak before rolling back to match $H$'s final certified predicate configuration. Historical evaluators suffer score inflation (1.0 vs. 0.5), while current-state evaluators remain invariant (0.5).}
\label{fig:historical-witness}
\end{figure}

Proposition 1 checks protocol integrity. The empirical question is whether legal witnesses can produce the failure described by Proposition 2. Figure~\ref{fig:historical-witness} illustrates this mechanism.

\subsection{Strict rollback}\label{strict-rollback}

A strict rollback starts from (H), preserves the active goal, and executes a legal inverse transition that makes at least one persistent predicate false. The certifier requires $P_{\mathrm{state}}(R)<P_{\mathrm{state}}(H)$ and verifies event order and goal version. Detection is $\mathbf{1}[f(R)<f(H)]$. Four GoalPatch tasks are excluded from this metric.

\subsection{Public/private isolation}\label{publicprivate-isolation}

The private package contains expected values, hidden tests, trusted state, private identifiers, certification vectors, and eligibility verdicts. A closed serializer produces a versioned public payload with the task instruction, public goal schema, redacted state, state diff, provenance, and transcript. Evaluators receive byte-identical payloads in public-only processes. Capability probes cover private reads, subprocesses, links, and path traversal.

IDJ receives a schema-free projection that removes the predicate schema and Oracle-derived fields. It generates one checklist from each task instruction and freezes it before scoring. Both the projection and checklist are versioned and hashed.

\subsection{Reference and evaluator-targeted adversaries}\label{reference-and-evaluator-targeted-adversaries}

The Historical Milestone run produces the shared set $A_{\mathrm{ref}}$, which supports paired transfer comparisons. Target-specific exploitability uses

\begin{equation}
A_f=\arg\max_{A\in\mathcal{E}_i}f(A).
\end{equation}

where $\mathcal{E}_i$ contains valid component-wise matched candidates. Each target receives ($K=3$) attempts. Selection uses the frozen public score and a deterministic tie-break after private eligibility checks.

Historical Milestone and Predicate-Max share one run because their outputs are identical. CAPE-Max, DeepSeek, and IDJ receive separate runs. CSPS and Lightweight CAPE require no search because their scores are exact deterministic functions of the corresponding certified components in PB-CSTE. They therefore serve as analytic positive controls.

\section{Experimental setup}\label{experimental-setup-1}

\subsection{Protocol and evaluators}\label{protocol-and-evaluators}

The held-out split, interface, prompts, model settings, retry rule, and ($K=3$) historical-target budget were sealed before scoring. The attacker generated public white-box stress candidates using deepseek-v4-pro in thinking mode, temperature 0, strict JSON, a 12,000-token limit, a 600-second timeout, and one retry for transport or schema failure. Among 54 attempts, 37 were protocol-valid and 15 tasks produced $A_{\mathrm{ref}}$. No matched adversary was found for three tasks.

Historical Milestone and Predicate-Max retain the highest observed completion fraction. CAPE-Max also applies standardized attribution. CSPS scores predicate satisfaction in the final public state, while Lightweight CAPE combines final-state satisfaction with attribution. The DeepSeek judge assigns a structured partial-completion score. Terminal Outcome reports binary completion (Table~\ref{tab:evaluator-positioning}).

IDJ uses Gemini 2.5 Pro with temperature 0, strict JSON, a 4,096-token limit, a 600-second timeout, and one bounded retry. It generates a schema-free checklist for each task and freezes the checklist before scoring. Each trajectory receives three scores, with the median used as the primary result. A prospectively frozen extension applies the same attacker and ($K=3$) budget separately to CAPE-Max, DeepSeek, and IDJ.

\begin{table}[htbp]
\centering
\caption{Evaluator positioning. IDJ sits outside the \(2 \times 2\) because independent model-based decomposition introduces another factor.}\label{tab:evaluator-positioning}
\small
\begin{tabular}{@{}
  >{\raggedright\arraybackslash}p{(\linewidth - 6\tabcolsep) * \real{0.2800}}
  >{\raggedright\arraybackslash}p{(\linewidth - 6\tabcolsep) * \real{0.1800}}
  >{\raggedright\arraybackslash}p{(\linewidth - 6\tabcolsep) * \real{0.1800}}
  >{\raggedright\arraybackslash}p{(\linewidth - 6\tabcolsep) * \real{0.3600}}@{}}
\toprule\noalign{}
\begin{minipage}[b]{\linewidth}\raggedright
\textbf{Evaluator}
\end{minipage} & \begin{minipage}[b]{\linewidth}\raggedright
\textbf{Time}
\end{minipage} & \begin{minipage}[b]{\linewidth}\raggedright
\textbf{Attribution}
\end{minipage} & \begin{minipage}[b]{\linewidth}\raggedright
\textbf{Decomposition}
\end{minipage} \\
\midrule\noalign{}
Predicate-Max & historical & no & shared \\
CAPE-Max & historical & yes & shared \\
CSPS & current & no & shared \\
Lightweight CAPE & current & yes & shared \\
DeepSeek judge & semantic & prompted & public schema \\
IDJ & semantic & prompted & independent \\
Terminal & terminal & no & completion bit \\
\bottomrule
\end{tabular}
\end{table}

\subsection{Metrics and denominators}\label{metrics-and-denominators}

The analysis addresses four questions: witness constructability, inflation magnitude, transfer to semantic judges, and sensitivity to strict rollbacks.

Honest accuracy is the MAE against $P_{\mathrm{state}}$ across all 18 honest trajectories; $P_{\mathrm{state}}=P_{\mathrm{agent}}$ holds on this frozen honest set.

\begin{equation*}
\begin{aligned}
\mathrm{coverage} &= \frac{n_{\mathrm{matched}}}{18}, &
\mathrm{conditional\ ASR} &= \frac{n_{\Delta_{\mathrm{hack}}>.10}}{n_{\mathrm{matched}}}, \\
\mathrm{end\text{-}to\text{-}end\ yield} &= \frac{n_{\Delta_{\mathrm{hack}}>.10}}{18}.
\end{aligned}
\end{equation*}

Mean $\Delta_{\mathrm{hack}}$ and its interval use matched tasks. Coverage and end-to-end yield retain generation failures in the denominator. Shared-$A_{\mathrm{ref}}$ results measure transfer, while $A_f$ results measure target-specific exploitability. The two sets are reported separately.

Rollback evaluation uses 14 fixed-goal pairs. The task is the statistical unit. We use 10,000-resample task or paired percentile bootstrap intervals and Wilson intervals for proportions. The frozen H1--H4 test mapping was underspecified, so the analysis remains exploratory and reports no post hoc Holm-controlled family.

\subsection{Integrity audits}\label{integrity-audits}

All serializer, leakage, no-verdict, same-payload, and private-access audits passed before scoring. Public requests were exactly reconstructable, and 197 private canaries produced zero hits. All 51 original and 396 extended judge calls returned valid structured outputs. A clean reproduction rebuilt the 51 trajectories and M5 metrics. An independent certifier agreed on all 51 cases, and ten mutation controls produced the expected results. These audits support the evidence chain; ecological validity remains outside their scope.

\section{Results}\label{results}

\subsection{Constructability and magnitude of historical witnesses}\label{constructability-and-magnitude-of-historical-witnesses}

The frozen Historical Milestone run produced matched adversaries for 15/18 tasks. Although every pair matches component-wise on $c_{\mathrm{state}}$ and $c_{\mathrm{agent}}$, adversaries receive .252 more credit on average (95\% task-bootstrap CI {[}.154, .351{]}). Ten pairs exceed the .10 threshold, giving a conditional ASR of 10/15 and an end-to-end yield of 10/18. Predicate-Max is output-identical on all 51 original trajectories and forms a degenerate historical contrast. CAPE-Max produces the same transfer mean and conditional ASR on $A_{\mathrm{ref}}$.

The result extends beyond the .5 gap observed in the development pilot. Honest progress spans seven levels, and per-task inflation ranges from 0 to .6. These findings quantify exploitability under prompted attacks and make no claim about spontaneous attack frequency.

\subsection{Semantic judges are resistant, not immune}\label{semantic-judges-are-resistant-not-immune-1}

The structured DeepSeek judge achieves an honest MAE of .036. On $A_{\mathrm{ref}}$, its mean transfer inflation is .022 (95\% CI {[}0, .067{]}), with an ASR of 1/15. It detects 10/14 rollbacks. Role separation limits direct prompt contamination, while the attacker and judge share the deepseek-v4-pro backbone.

IDJ achieves an honest MAE of .041. Its mean transfer inflation is .017 (95\% CI {[}.003, .034{]}), with an ASR of 1/15. It detects 12/14 rollbacks and retains mean false credit of .018 across all pairs and .125 across the two misses. The median within-trajectory standard deviation across three calls is .009. A blind alignment audit identifies 49/79 checklist conditions as one-to-one, 21/79 as merged, split, or indirect, and 9/79 as absent or implicit.

IDJ extends the evidence beyond Oracle-aligned decomposition. Its shared-set results measure transfer under a common task intent.

\subsection{Current-state rules satisfy the expected invariance}\label{current-state-rules-satisfy-the-expected-invariance}

As guaranteed by Proposition 1, CSPS and Lightweight CAPE yield $\Delta_{\mathrm{hack}}=0$ on every matched pair. This result verifies the structural control. Both evaluators also achieve zero honest MAE. Terminal Outcome yields zero inflation with an MAE of .540, showing the information loss caused by binary scoring.

\subsection{Evaluator-targeted attacks}\label{evaluator-targeted-attacks-1}

Target-specific runs measure exploitability under direct optimization. Direct targeting increases inflation for both semantic judges. Historical targets retain the highest end-to-end yields, followed by DeepSeek and IDJ. Conditional means use each target's matched task set (Table~\ref{tab:targeted-results}).

\begin{table}[htbp]
\centering
\caption{Target-specific attack results. Mean \(\Delta_{\mathrm{hack}}\) is conditional on matched tasks.}\label{tab:targeted-results}
\small
\begin{tabular}{@{}
  >{\raggedright\arraybackslash}p{(\linewidth - 10\tabcolsep) * \real{0.2600}}
  >{\raggedright\arraybackslash}p{(\linewidth - 10\tabcolsep) * \real{0.1000}}
  >{\raggedright\arraybackslash}p{(\linewidth - 10\tabcolsep) * \real{0.1300}}
  >{\raggedright\arraybackslash}p{(\linewidth - 10\tabcolsep) * \real{0.2400}}
  >{\raggedright\arraybackslash}p{(\linewidth - 10\tabcolsep) * \real{0.1300}}
  >{\raggedright\arraybackslash}p{(\linewidth - 10\tabcolsep) * \real{0.1400}}@{}}
\toprule\noalign{}
\begin{minipage}[b]{\linewidth}\raggedright
\textbf{Target}
\end{minipage} & \begin{minipage}[b]{\linewidth}\raggedright
\textbf{Valid}
\end{minipage} & \begin{minipage}[b]{\linewidth}\raggedright
\textbf{Coverage}
\end{minipage} & \begin{minipage}[b]{\linewidth}\raggedright
\textbf{Mean} \(\Delta_{\mathrm{hack}}\) \textbf{{[}95\% CI{]}}
\end{minipage} & \begin{minipage}[b]{\linewidth}\raggedright
\textbf{Cond. ASR}
\end{minipage} & \begin{minipage}[b]{\linewidth}\raggedright
\textbf{E2E yield}
\end{minipage} \\
\midrule\noalign{}
Historical/\allowbreak Predicate-Max & 37/54 & 15/18 & .252 {[}.154, .351{]} & 10/15 & 10/18 \\
CAPE-Max & 39/54 & 16/18 & .239 {[}.150, .335{]} & 10/16 & 10/18 \\
DeepSeek judge & 34/54 & 14/18 & .069 {[}.031, .113{]} & 4/14 & 4/18 \\
IDJ & 36/54 & 15/18 & .048 {[}.024, .074{]} & 3/15 & 3/18 \\
CSPS / Lightweight CAPE & --- & analytic & identically 0 & --- & --- \\
\bottomrule
\end{tabular}
\end{table}

\subsection{Rollback exposes the same temporal failure}\label{rollback-exposes-the-same-temporal-failure}

By construction, historical evaluators detect $0/14$ strict rollbacks and retain mean false credit of $.448$--$.504$ across all pairs, while CSPS and Lightweight CAPE detect $14/14$ and retain zero false credit. These outcomes are analytic sanity checks implied by the evaluator definitions and the rollback eligibility rule. DeepSeek detects $10/14$
rollbacks and retains $.123$ false credit across four misses. IDJ detects $12/14$ and retains $.125$ false credit across two misses. The semantic-judge results provide the empirical evidence on rollback sensitivity.

\subsection{Provenance has a narrow attribution benefit}\label{provenance-has-a-narrow-attribution-benefit}

Under the tested attribution rule, CSPS and Lightweight CAPE have identical honest MAE, shared-set inflation, and rollback detection. The observed robustness gains are therefore associated with current-state accounting (Table~\ref{tab:shared-results}). Evidence about provenance remains limited to the four designated attribution cases.

\begin{table}[htbp]
\centering
\caption{Core results on the shared reference set.}\label{tab:shared-results}
\small
\begin{tabular}{@{}
  >{\raggedright\arraybackslash}p{(\linewidth - 6\tabcolsep) * \real{0.4000}}
  >{\raggedright\arraybackslash}p{(\linewidth - 6\tabcolsep) * \real{0.2000}}
  >{\raggedright\arraybackslash}p{(\linewidth - 6\tabcolsep) * \real{0.2000}}
  >{\raggedright\arraybackslash}p{(\linewidth - 6\tabcolsep) * \real{0.2000}}@{}}
\toprule\noalign{}
\begin{minipage}[b]{\linewidth}\raggedright
\textbf{Evaluator}
\end{minipage} & \begin{minipage}[b]{\linewidth}\raggedright
\textbf{Honest MAE}
\end{minipage} & \begin{minipage}[b]{\linewidth}\raggedright
\textbf{Mean} \(\Delta_{\mathrm{hack}}\)
\end{minipage} & \begin{minipage}[b]{\linewidth}\raggedright
\textbf{Rollback}
\end{minipage} \\
\midrule\noalign{}
Historical / Predicate & 0 & .252 & 0/14 \\
CAPE-Max & 0 & .252 & 0/14 \\
DeepSeek judge & .036 & .022 & 10/14 \\
CSPS & 0 & 0 & 14/14 \\
Lightweight CAPE & 0 & 0 & 14/14 \\
IDJ & .041 & .017 & 12/14 \\
Terminal Outcome & .540 & 0 & auxiliary \\
\bottomrule
\end{tabular}
\end{table}

\section{Mechanism analysis and discussion}\label{mechanism-analysis-and-discussion}

\subsection{Temporal accounting explains the main pattern}\label{temporal-accounting-explains-the-main-pattern}

The controlled $2\times2$ varies temporal accounting and provenance. On the same $A_{\mathrm{ref}}$ trajectories, replacing Predicate-Max with CSPS reduces adversarial false credit by .252 (95\% bootstrap CI {[}.154, .351{]}) and rollback false credit by .504 ({[}.388, .629{]}). Replacing CAPE-Max with Lightweight CAPE produces similar reductions: .252 for matched attacks and .448 for rollbacks. Provenance does not change these outcomes in the tested fixtures. The paired results therefore associate the robustness gains with temporal accounting.

IDJ and DeepSeek show the same general pattern. Both are more resistant than the historical rules, although residual failures remain. Semantic inference therefore introduces additional vulnerabilities beyond temporal accounting.

\subsection{Oracle--CSPS qualification}\label{oraclecsps-qualification}

The private certifier and CSPS are physically isolated. In PB-CSTE, each ordered public predicate evaluated by CSPS has a fixed one-to-one correspondence with a component of $c_{\mathrm{state}}$, while Lightweight CAPE additionally uses the corresponding component of $c_{\mathrm{agent}}$. Hence their scores are exact deterministic functions of the certified vectors and satisfy Proposition~1. This construction makes them structural positive controls.

This result supports a narrow claim: aligned current-state semantics eliminate the observed PB-CSTE attacks. IDJ provides complementary evidence under a less aligned decomposition. It receives no Oracle predicate schema and constructs a frozen checklist from the task instruction using a different model family. Some semantic correlation may remain because both systems interpret the same instruction.

\subsection{Evaluator robustness as a first-class object}\label{evaluator-robustness-as-a-first-class-object}

PartHackBench evaluates the mapping from behavior to score. Equal-progress pairs test score invariance when trajectories preserve the certified goal configuration. Rollback pairs test sensitivity to certified decreases. Shared-reference attacks support paired mechanism comparisons, while evaluator-targeted attacks measure prompted exploitability. Separate reporting prevents transfer failure, generation failure, and loss of partial-progress information from being interpreted as robustness.

The findings support a hybrid design. Persistent conditions can be grounded in trusted current state. Standardized provenance can assign actor credit. Semantic judges can interpret ambiguous evidence. Each component should be tested against certified invariances. Benchmarks should also define persistence semantics, include rollback controls, and isolate certification from the evaluator.

\section{Limitations and conclusion}\label{limitations-and-conclusion}

As an initial study, PartHackBench contains 24 tasks in total (6 development and 18 held-out); future work will expand the task families, environments, and evaluator coverage to improve statistical power and test generalization in realistic settings.

PB-CSTE omits ecological noise to isolate evaluator vulnerabilities. Its 18 held-out tasks form a compact controlled test set, which limits external validity. Bootstrap intervals summarize uncertainty across these tasks. Directed white-box attacks measure prompted exploitability, not deployment prevalence. Conditional ASR covers tasks with matched candidates, while end-to-end yield also includes generation failure. Mean $\Delta_{\mathrm{hack}}$ remains conditional on matched tasks.

Target-specific search uses ($K=3$) attempts per task. Historical Milestone and Predicate-Max are output-identical. Provenance evidence covers four designated cases. IDJ uses an independently generated decomposition, while shared task instructions may still create semantic alignment. The certification guarantee covers encoded predicates and standardized attribution. It excludes trajectory cost, length, language quality, and task-external effects.

Within these limits, PartHackBench provides a controlled test for partial-credit evaluators. Historical accounting inflates scores at matched final progress and retains stale credit after rollbacks. Semantic judges reduce these failures but do not eliminate them. Current-state controls satisfy the expected invariance. Evaluators should therefore be tested under independently certified transformations of the task progress they claim to measure.

\clearpage
\begingroup\small
\bibliographystyle{plainnat}
\bibliography{references}

@inproceedings{yao2025taubench,title={{${\tau}$-bench}: A Benchmark for Tool-Agent-User Interaction in Real-World Domains},author={Yao, Shunyu and Shinn, Noah and Razavi, Pedram and Narasimhan, Karthik},booktitle={ICLR},year={2025},url={https://proceedings.iclr.cc/paper_files/paper/2025/hash/1b126cc38b8638e07bef37e7b2bb72bf-Abstract-Conference.html}}

@inproceedings{barres2026tau2,title={{${\tau^2}$-Bench}: Evaluating Conversational Agents in a Dual-Control Environment},author={Barres, Victor and Dong, Honghua and Ray, Soham and Si, Xujie and Narasimhan, Karthik},booktitle={ICML},series={PMLR},volume={306},year={2026},url={https://openreview.net/forum?id=OC2z7iSQKa}}

@inproceedings{lu2025toolsandbox,title={{ToolSandbox}: A Stateful, Conversational, Interactive Evaluation Benchmark for {LLM} Tool Use Capabilities},author={Lu, Jiarui and Holleis, Thomas and Zhang, Yizhe and Aumayer, Bernhard and Nan, Feng and Bai, Haoping and Ma, Shuang and Ma, Shen and Li, Mengyu and Yin, Guoli and Wang, Zirui and Pang, Ruoming},booktitle={Findings of NAACL},pages={1160--1183},year={2025},doi={10.18653/v1/2025.findings-naacl.65}}

@inproceedings{jimenez2024swebench,title={{SWE}-bench: Can Language Models Resolve Real-World {GitHub} Issues?},author={Jimenez, Carlos E. and Yang, John and Wettig, Alexander and Yao, Shunyu and Pei, Kexin and Press, Ofir and Narasimhan, Karthik},booktitle={ICLR},year={2024},url={https://proceedings.iclr.cc/paper_files/paper/2024/hash/edac78c3e300629acfe6cbe9ca88fb84-Abstract-Conference.html}}

@inproceedings{zhou2024webarena,title={{WebArena}: A Realistic Web Environment for Building Autonomous Agents},author={Zhou, Shuyan and Xu, Frank F. and Zhu, Hao and Zhou, Xuhui and Lo, Robert and Sridhar, Abishek and Cheng, Xianyi and Ou, Tianyue and Bisk, Yonatan and Fried, Daniel and Alon, Uri and Neubig, Graham},booktitle={ICLR},year={2024},url={https://proceedings.iclr.cc/paper_files/paper/2024/hash/4410c0711e9154a7a2d26f9b3816d1ef-Abstract-Conference.html}}

@inproceedings{drouin2024workarena,title={{WorkArena}: How Capable are Web Agents at Solving Common Knowledge Work Tasks?},author={Drouin, Alexandre and Gasse, Maxime and Caccia, Massimo and Laradji, Issam H. and Del Verme, Manuel and Marty, Tom and Vazquez, David and Chapados, Nicolas and Lacoste, Alexandre},booktitle={ICML},series={PMLR},volume={235},pages={11642--11662},year={2024},url={https://proceedings.mlr.press/v235/drouin24a.html}}

@inproceedings{boisvert2024workarenapp,title={{WorkArena++}: Towards Compositional Planning and Reasoning-based Common Knowledge Work Tasks},author={Boisvert, L{\'e}o and Thakkar, Megh and Gasse, Maxime and Caccia, Massimo and Le Sellier de Chezelles, Thibault and Cappart, Quentin and Chapados, Nicolas and Lacoste, Alexandre and Drouin, Alexandre},booktitle={NeurIPS Datasets and Benchmarks},year={2024},doi={10.52202/079017-0195}}

@inproceedings{liu2024agentbench,title={{AgentBench}: Evaluating {LLM}s as Agents},author={Liu, Xiao and Yu, Hao and Zhang, Hanchen and Xu, Yifan and Lei, Xuanyu and Lai, Hanyu and Gu, Yu and Ding, Hangliang and Men, Kaiwen and Yang, Kejuan and Zhang, Shudan and Deng, Xiang and Zeng, Aohan and Du, Zhengxiao and Zhang, Chenhui and Shen, Sheng and Zhang, Tianjun and Su, Yu and Sun, Huan and Huang, Minlie and Dong, Yuxiao and Tang, Jie},booktitle={ICLR},year={2024},url={https://proceedings.iclr.cc/paper_files/paper/2024/hash/e9df36b21ff4ee211a8b71ee8b7e9f57-Abstract-Conference.html}}

@inproceedings{li2023apibank,title={{API}-Bank: A Comprehensive Benchmark for Tool-Augmented {LLM}s},author={Li, Minghao and Zhao, Yingxiu and Yu, Bowen and Song, Feifan and Li, Hangyu and Yu, Haiyang and Li, Zhoujun and Huang, Fei and Li, Yongbin},booktitle={EMNLP},pages={3102--3116},year={2023},doi={10.18653/v1/2023.emnlp-main.187}}

@inproceedings{patil2025bfcl,title={The Berkeley Function Calling Leaderboard (BFCL): From Tool Use to Agentic Evaluation of Large Language Models},author={Patil, Shishir G. and Mao, Huanzhi and Yan, Fanjia and Ji, Charlie Cheng-Jie and Suresh, Vishnu and Stoica, Ion and Gonzalez, Joseph E.},booktitle={ICML},series={PMLR},volume={267},pages={48371--48392},year={2025},url={https://proceedings.mlr.press/v267/patil25a.html}}

@inproceedings{ma2024agentboard,title={{AgentBoard}: An Analytical Evaluation Board of Multi-turn {LLM} Agents},author={Ma, Chang and Zhang, Junlei and Zhu, Zhihao and Yang, Cheng and Yang, Yujiu and Jin, Yaohui and Lan, Zhenzhong and Kong, Lingpeng and He, Junxian},booktitle={NeurIPS Datasets and Benchmarks},volume={37},pages={74325--74362},year={2024},doi={10.52202/079017-2365}}

@inproceedings{lu2025agentrewardbench,title={{AgentRewardBench}: Evaluating Automatic Evaluations of Web Agent Trajectories},author={L{\`u}, Xing Han and Kazemnejad, Amirhossein and Meade, Nicholas and Patel, Arkil and Shin, Dongchan and Zambrano, Alejandra and Sta{\'n}czak, Karolina and Shaw, Peter and Pal, Christopher J. and Reddy, Siva},booktitle={COLM},year={2025},eprint={2504.08942},url={https://openreview.net/forum?id=fQcUZMPIvu}}

@inproceedings{li2026toolprmbench,title={{ToolPRMBench}: Evaluating and Advancing Process Reward Models for Tool-using Agents},author={Li, Dawei and Yao, Yuguang and Tan, Zhen and Liu, Huan and Guo, Ruocheng},booktitle={Findings of ACL},pages={12378--12391},year={2026},doi={10.18653/v1/2026.findings-acl.602}}

@inproceedings{lightman2024verify,title={Let's Verify Step by Step},author={Lightman, Hunter and Kosaraju, Vineet and Burda, Yuri and Edwards, Harrison and Baker, Bowen and Lee, Teddy and Leike, Jan and Schulman, John and Sutskever, Ilya and Cobbe, Karl},booktitle={ICLR},year={2024},url={https://proceedings.iclr.cc/paper_files/paper/2024/hash/aca97732e30bcf1303bc22ac3924fd16-Abstract-Conference.html}}

@article{uesato2022processoutcome,title={Solving Math Word Problems with Process- and Outcome-Based Feedback},author={Uesato, Jonathan and Kushman, Nate and Kumar, Ramana and Song, Francis and Siegel, Noah and Wang, Lisa and Creswell, Antonia and Irving, Geoffrey and Higgins, Irina},journal={arXiv preprint arXiv:2211.14275},year={2022},eprint={2211.14275},doi={10.48550/arXiv.2211.14275}}

@inproceedings{zheng2023mtbench,title={Judging {LLM}-as-a-Judge with {MT}-Bench and Chatbot Arena},author={Zheng, Lianmin and Chiang, Wei-Lin and Sheng, Ying and Zhuang, Siyuan and Wu, Zhanghao and Zhuang, Yonghao and Lin, Zi and Li, Zhuohan and Li, Dacheng and Xing, Eric P. and Zhang, Hao and Gonzalez, Joseph E. and Stoica, Ion},booktitle={NeurIPS Datasets and Benchmarks},year={2023},doi={10.52202/075280-2020}}

@inproceedings{liu2023geval,title={{G}-Eval: {NLG} Evaluation using {GPT}-4 with Better Human Alignment},author={Liu, Yang and Iter, Dan and Xu, Yichong and Wang, Shuohang and Xu, Ruochen and Zhu, Chenguang},booktitle={EMNLP},pages={2511--2522},year={2023},doi={10.18653/v1/2023.emnlp-main.153}}

@inproceedings{kim2024prometheus,title={{Prometheus}: Inducing Fine-Grained Evaluation Capability in Language Models},author={Kim, Seungone and Shin, Jamin and Cho, Yejin and Jang, Joel and Longpre, Shayne and Lee, Hwaran and Yun, Sangdoo and Shin, Seongjin and Kim, Sungdong and Thorne, James and Seo, Minjoon},booktitle={ICLR},year={2024},url={https://proceedings.iclr.cc/paper_files/paper/2024/hash/803485352e61e3ebf41221e4776c9fd4-Abstract-Conference.html}}

@inproceedings{wang2024unfair,title={Large Language Models are not Fair Evaluators},author={Wang, Peiyi and Li, Lei and Chen, Liang and Cai, Zefan and Zhu, Dawei and Lin, Binghuai and Cao, Yunbo and Kong, Lingpeng and Liu, Qi and Liu, Tianyu and Sui, Zhifang},booktitle={ACL},pages={9440--9450},year={2024},doi={10.18653/v1/2024.acl-long.511}}

@inproceedings{panickssery2024selfpreference,title={{LLM} Evaluators Recognize and Favor Their Own Generations},author={Panickssery, Arjun and Bowman, Samuel R. and Feng, Shi},booktitle={NeurIPS},year={2024},doi={10.52202/079017-2197}}

@inproceedings{tan2025judgebench,title={{JudgeBench}: A Benchmark for Evaluating {LLM}-Based Judges},author={Tan, Sijun and Zhuang, Siyuan and Montgomery, Kyle and Tang, William Y. and Cuadron, Alejandro and Wang, Chenguang and Popa, Raluca Ada and Stoica, Ion},booktitle={ICLR},year={2025},url={https://proceedings.iclr.cc/paper_files/paper/2025/hash/9e720fce64f91114c49cfd640d821da3-Abstract-Conference.html}}

@inproceedings{li2025calibraeval,title={{CalibraEval}: Calibrating Prediction Distribution to Mitigate Selection Bias in {LLM}s-as-Judges},author={Li, Haitao and Chen, Junjie and Ai, Qingyao and Chu, Zhumin and Zhou, Yujia and Dong, Qian and Liu, Yiqun},booktitle={ACL},pages={16537--16552},year={2025},doi={10.18653/v1/2025.acl-long.808}}

@inproceedings{lambert2025rewardbench,title={{RewardBench}: Evaluating Reward Models for Language Modeling},author={Lambert, Nathan and Pyatkin, Valentina and Morrison, Jacob and Miranda, LJ and Lin, Bill Yuchen and Chandu, Khyathi and Dziri, Nouha and Kumar, Sachin and Zick, Tom and Choi, Yejin and Smith, Noah A. and Hajishirzi, Hannaneh},booktitle={Findings of NAACL},pages={1755--1797},year={2025},doi={10.18653/v1/2025.findings-naacl.96}}

@article{amodei2016concrete,title={Concrete Problems in {AI} Safety},author={Amodei, Dario and Olah, Chris and Steinhardt, Jacob and Christiano, Paul and Schulman, John and Man{\'e}, Dan},journal={arXiv preprint arXiv:1606.06565},year={2016},eprint={1606.06565},doi={10.48550/arXiv.1606.06565}}

@inproceedings{skalse2022rewardgaming,title={Defining and Characterizing Reward Gaming},author={Skalse, Joar Max Viktor and Howe, Nikolaus H. R. and Krasheninnikov, Dmitrii and Krueger, David},booktitle={NeurIPS},volume={35},pages={9460--9471},year={2022},doi={10.52202/068431-0687}}

@inproceedings{gao2023overoptimization,title={Scaling Laws for Reward Model Overoptimization},author={Gao, Leo and Schulman, John and Hilton, Jacob},booktitle={ICML},series={PMLR},volume={202},pages={10835--10866},year={2023},url={https://proceedings.mlr.press/v202/gao23h.html}}

@techreport{w3c2013prov,title={{PROV-DM}: The {PROV} Data Model},author={{World Wide Web Consortium}},editor={Moreau, Luc and Missier, Paolo},institution={World Wide Web Consortium},type={{W3C} Recommendation},year={2013},url={https://www.w3.org/TR/2013/REC-prov-dm-20130430/}}

@inproceedings{chuang2026proxy,title={Toward Scalable Verifiable Reward: Proxy State-Based Evaluation for Multi-turn Tool-Calling {LLM} Agents},author={Chuang, Yun-Shiuan and Kulkarni, Chaitanya and Chiu, Alec M. and Thangali, Avinash and Pan, Zijie and Shekhar, Shivani and Ge, Yirou and Li, Yixi and Kona, Uma and Pang, Linsey and Mehrotra, Prakhar},booktitle={ACL Industry Track},pages={1251--1264},year={2026},doi={10.18653/v1/2026.acl-industry.87}}
\endgroup

\clearpage
\appendix
\counterwithin{table}{section}
\counterwithin{figure}{section}
\counterwithin{equation}{section}
\section{Benchmark and implementation details}\label{a.-benchmark-and-implementation-details}

\Needspace{6\baselineskip}
\subsection{PB-CSTE and task construction}\label{a.1-pb-cste-and-task-construction}

PB-CSTE v1.0 is a project-local deterministic environment for causal isolation in stateful tool use. Tasks manipulate typed entities through allowlisted create, update, delete, restore, and goal-revision operations. Every successful operation records a timestamp, wrapper-assigned actor class, state diff, and post-action snapshot in an append-only log; failed actions do not mutate state. Replay begins from a frozen initial snapshot, and state equality is exact over typed values. The setting intentionally omits networks, concurrency, permission systems, stochastic tool failure, and open browsing.

The frozen benchmark contains 6 development and 18 held-out tasks. Each task has three to five persistent, independently testable goal predicates. Family counts are mutually exclusive; attribution and rollback tags overlap families.

\begingroup\small
\begin{longtable}[]{@{}
  >{\raggedright\arraybackslash}p{(\linewidth - 6\tabcolsep) * \real{0.3859}}
  >{\raggedleft\arraybackslash}p{(\linewidth - 6\tabcolsep) * \real{0.1978}}
  >{\raggedleft\arraybackslash}p{(\linewidth - 6\tabcolsep) * \real{0.1978}}
  >{\raggedleft\arraybackslash}p{(\linewidth - 6\tabcolsep) * \real{0.1978}}@{}}
\caption{Task-family allocation.}\label{tab:task-families}\tabularnewline
\toprule\noalign{}
\begin{minipage}[b]{\linewidth}\raggedright
\textbf{Family}
\end{minipage} & \begin{minipage}[b]{\linewidth}\raggedleft
\textbf{Development}
\end{minipage} & \begin{minipage}[b]{\linewidth}\raggedleft
\textbf{Held-out}
\end{minipage} & \begin{minipage}[b]{\linewidth}\raggedleft
\textbf{Total}
\end{minipage} \\
\midrule\noalign{}
\endfirsthead
\toprule\noalign{}
\begin{minipage}[b]{\linewidth}\raggedright
\textbf{Family}
\end{minipage} & \begin{minipage}[b]{\linewidth}\raggedleft
\textbf{Development}
\end{minipage} & \begin{minipage}[b]{\linewidth}\raggedleft
\textbf{Held-out}
\end{minipage} & \begin{minipage}[b]{\linewidth}\raggedleft
\textbf{Total}
\end{minipage} \\
\midrule\noalign{}
\endhead
\bottomrule\noalign{}
\endlastfoot
Persistent multi-step & 2 & 6 & 8 \\
Entity CRUD & 1 & 5 & 6 \\
Fixed-goal interaction & 1 & 3 & 4 \\
GoalPatch stress & 2 & 4 & 6 \\
\textbf{Total} & \textbf{6} & \textbf{18} & \textbf{24} \\
\end{longtable}
\endgroup

Honest trajectories are legal partial prefixes with no terminal success. Held-out progress spans \(1/4,1/3,2/5,1/2,3/5,2/3,\) and \(3/4\), avoiding the fixed-\(0.5\) artifact of the development pilot. GoalPatch tasks bind certification to the active goal version. Replays and evaluator calls use separate project-local processes; tool execution has no network access, and only the certifier mounts the private task package.

\Needspace{6\baselineskip}
\subsection{Held-out task manifest}\label{a.2-held-out-task-manifest}

Table~\ref{tab:heldout-manifest} reports public structural metadata only. Expected values, hidden tests, private identifiers, and certified vectors remain sealed. ``Attr.'' and ``RB'' mark four attribution-sensitive and four designated rollback-sensitive fixtures. All 14 fixed-goal tasks admit strict rollback construction; the four GoalPatch tasks are excluded from that metric.

\begingroup\small
\begin{longtable}[]{@{}
  >{\raggedright\arraybackslash}p{(\linewidth - 14\tabcolsep) * \real{0.1007}}
  >{\raggedright\arraybackslash}p{(\linewidth - 14\tabcolsep) * \real{0.1501}}
  >{\raggedright\arraybackslash}p{(\linewidth - 14\tabcolsep) * \real{0.1278}}
  >{\raggedleft\arraybackslash}p{(\linewidth - 14\tabcolsep) * \real{0.0504}}
  >{\centering\arraybackslash}p{(\linewidth - 14\tabcolsep) * \real{0.1259}}
  >{\centering\arraybackslash}p{(\linewidth - 14\tabcolsep) * \real{0.0697}}
  >{\centering\arraybackslash}p{(\linewidth - 14\tabcolsep) * \real{0.0920}}
  >{\centering\arraybackslash}p{(\linewidth - 14\tabcolsep) * \real{0.1695}}@{}}
\caption{Held-out manifest and frozen honest progress.}\label{tab:heldout-manifest}\tabularnewline
\toprule\noalign{}
\begin{minipage}[b]{\linewidth}\raggedright
\textbf{ID}
\end{minipage} & \begin{minipage}[b]{\linewidth}\raggedright
\textbf{Family}
\end{minipage} & \begin{minipage}[b]{\linewidth}\raggedright
\textbf{Goal form}
\end{minipage} & \begin{minipage}[b]{\linewidth}\raggedleft
\(m\)
\end{minipage} & \begin{minipage}[b]{\linewidth}\centering
\(P_{\mathrm{state}}(H)\)
\end{minipage} & \begin{minipage}[b]{\linewidth}\centering
\textbf{Tag}
\end{minipage} & \begin{minipage}[b]{\linewidth}\centering
\(A_{\mathrm{ref}}\)
\end{minipage} & \begin{minipage}[b]{\linewidth}\centering
\textbf{RB metric}
\end{minipage} \\
\midrule\noalign{}
\endfirsthead
\toprule\noalign{}
\begin{minipage}[b]{\linewidth}\raggedright
\textbf{ID}
\end{minipage} & \begin{minipage}[b]{\linewidth}\raggedright
\textbf{Family}
\end{minipage} & \begin{minipage}[b]{\linewidth}\raggedright
\textbf{Goal form}
\end{minipage} & \begin{minipage}[b]{\linewidth}\raggedleft
\(m\)
\end{minipage} & \begin{minipage}[b]{\linewidth}\centering
\(P_{\mathrm{state}}(H)\)
\end{minipage} & \begin{minipage}[b]{\linewidth}\centering
\textbf{Tag}
\end{minipage} & \begin{minipage}[b]{\linewidth}\centering
\(A_{\mathrm{ref}}\)
\end{minipage} & \begin{minipage}[b]{\linewidth}\centering
\textbf{RB metric}
\end{minipage} \\
\midrule\noalign{}
\endhead
\bottomrule\noalign{}
\endlastfoot
PH-H01 & GoalPatch & revision & 4 & 1/4 & - & yes & excluded \\
PH-H02 & GoalPatch & revision & 3 & 1/3 & - & yes & excluded \\
PH-H03 & GoalPatch & revision & 5 & 2/5 & - & yes & excluded \\
PH-H04 & Persistent & multi-step & 5 & 2/5 & RB & yes & included \\
PH-H05 & Persistent & multi-step & 4 & 1/2 & Attr. & yes & included \\
PH-H06 & CRUD & entity set & 4 & 1/2 & - & yes & included \\
PH-H07 & GoalPatch & revision & 4 & 1/2 & - & yes & excluded \\
PH-H08 & Persistent & multi-step & 4 & 1/2 & - & yes & included \\
PH-H09 & CRUD & entity set & 4 & 1/2 & Attr. & yes & included \\
PH-H10 & Persistent & multi-step & 5 & 3/5 & RB & yes & included \\
PH-H11 & CRUD & entity set & 5 & 3/5 & - & yes & included \\
PH-H12 & Interaction & fixed goal & 5 & 3/5 & - & yes & included \\
PH-H13 & Persistent & multi-step & 5 & 3/5 & RB & yes & included \\
PH-H14 & CRUD & entity set & 5 & 3/5 & Attr. & yes & included \\
PH-H15 & Interaction & fixed goal & 3 & 2/3 & - & yes & included \\
PH-H16 & Persistent & multi-step & 3 & 2/3 & - & no & included \\
PH-H17 & CRUD & entity set & 4 & 3/4 & Attr. & no & included \\
PH-H18 & Interaction & fixed goal & 4 & 3/4 & RB & no & included \\
\end{longtable}
\endgroup

Each honest trajectory was checked for accidental transient peaks. A candidate is rejected for malformed schema, illegal action, replay mismatch, stale goal version, final-state vector mismatch, or attribution-vector mismatch. Generation failure remains in the 18-task yield denominator and is excluded from match-conditional effect estimates.

\Needspace{6\baselineskip}
\subsection{Certification and attribution}\label{a.3-certification-and-attribution}

For a trajectory \(\tau\) with trusted terminal state \(s_{T}\) and active predicates \(g_{1},\ldots,g_{m}\), the certifier constructs

\begin{equation}
\begin{aligned}
\mathbf{c}_{\mathrm{state}}(\tau)
  &= \Bigl(\mathbf{1}[g_j(s_T)=\mathrm{true}]\Bigr)_{j=1}^{m}, \\
\mathbf{c}_{\mathrm{agent}}(\tau)
  &= \Bigl(\mathbf{1}[g_j(s_T)=\mathrm{true}]\, \\
  &\qquad \mathbf{1}[\operatorname{latestCause}(g_j)=\mathrm{agent}]\Bigr)_{j=1}^{m}.
\end{aligned}
\end{equation}

For every satisfied predicate, the certifier traverses the trusted event log backward to the latest successful event that changed predicate-relevant state. The wrapper-assigned actor class supplies standardized credit; restore operations receive the restoring event's actor class. This is a reproducible event-log rule, not a claim of general causal attribution. Scalar summaries are

\begin{equation}
\begin{aligned}
P_{\mathrm{state}}(\tau) &= m^{-1}\mathbf{1}^{\top}\mathbf{c}_{\mathrm{state}}(\tau), &
P_{\mathrm{agent}}(\tau) &= m^{-1}\mathbf{1}^{\top}\mathbf{c}_{\mathrm{agent}}(\tau).
\end{aligned}
\end{equation}

An honest/adversarial pair is eligible only when, under the same active goal version,

\begin{equation}
\label{eq:appendix-matching}
\mathbf{c}_{\mathrm{state}}(H)=\mathbf{c}_{\mathrm{state}}(A),\qquad
\mathbf{c}_{\mathrm{agent}}(H)=\mathbf{c}_{\mathrm{agent}}(A).
\end{equation}

Matching is component-wise; scalar equality alone is insufficient. If \(f(\tau) = \phi\left( \mathbf{c}_{\mathrm{state}}(\tau),\mathbf{c}_{\mathrm{agent}}(\tau) \right)\), Eq.~\eqref{eq:appendix-matching} implies \(\Delta_{\mathrm{hack}} = f(A) - f(H) = 0\). Conversely, for \(f_{\mathrm{hist}}(\tau) = \max_{t \leq T}\psi\left( \mathbf{c}_{t} \right)\), a larger interior witness in \(A\) can increase the score while leaving both terminal vectors unchanged. These two observations give the matched-vector positive control and the historical-witness failure mechanism used in the main text.

The private package contains expected values, hidden tests, raw trusted state, private IDs, certified vectors, and eligibility verdicts. Original evaluators receive raw-byte-identical public payloads containing the task instruction, public goal schema, redacted terminal state and diff, standardized provenance, and transcript. IDJ receives a deterministic projection of the same record with the public predicate schema and Oracle-derived fields removed.

Two compact cases illustrate the gates. An eligible historical witness may satisfy \(\mathbf{c}_{\mathrm{state}}(H) = \mathbf{c}_{\mathrm{state}}(A) = \mathbf{c}_{\mathrm{agent}}(H) = \mathbf{c}_{\mathrm{agent}}(A) = (1,1,0)\), so both terminate at \(2/3\), while \(A\) temporarily reaches \((1,1,1)\) before a legal reversal. Historical credit remains \(1\) and current-state credit returns to \(2/3\). By contrast, equal state vectors \((1,1,0)\) with attribution vectors \((1,0,0)\) and \((0,1,0)\) have the same scalar average but fail Eq.~\eqref{eq:appendix-matching} and are rejected before scoring.

\Needspace{6\baselineskip}
\subsection{Evaluator configurations}\label{a.4-evaluator-configurations}

Table~\ref{tab:appendix-evaluators} places the evaluators by temporal rule and decomposition. Predicate-Max is output-identical to Historical Milestone in the original run. CSPS and Lightweight CAPE are deterministic functions of matched components and therefore serve as structurally aligned positive controls rather than independent robustness evidence.

\Needspace{14\baselineskip}
\begingroup\small
\begin{longtable}[]{@{}
  >{\raggedright\arraybackslash}p{(\linewidth - 6\tabcolsep) * \real{0.3183}}
  >{\raggedright\arraybackslash}p{(\linewidth - 6\tabcolsep) * \real{0.2122}}
  >{\raggedright\arraybackslash}p{(\linewidth - 6\tabcolsep) * \real{0.1736}}
  >{\raggedright\arraybackslash}p{(\linewidth - 6\tabcolsep) * \real{0.2749}}@{}}
\caption{Evaluator positioning.}\label{tab:appendix-evaluators}\tabularnewline
\toprule\noalign{}
\begin{minipage}[b]{\linewidth}\raggedright
\textbf{Evaluator}
\end{minipage} & \begin{minipage}[b]{\linewidth}\raggedright
\textbf{Temporal rule}
\end{minipage} & \begin{minipage}[b]{\linewidth}\raggedright
\textbf{Attribution}
\end{minipage} & \begin{minipage}[b]{\linewidth}\raggedright
\textbf{Decomposition}
\end{minipage} \\
\midrule\noalign{}
\endfirsthead
\toprule\noalign{}
\begin{minipage}[b]{\linewidth}\raggedright
\textbf{Evaluator}
\end{minipage} & \begin{minipage}[b]{\linewidth}\raggedright
\textbf{Temporal rule}
\end{minipage} & \begin{minipage}[b]{\linewidth}\raggedright
\textbf{Attribution}
\end{minipage} & \begin{minipage}[b]{\linewidth}\raggedright
\textbf{Decomposition}
\end{minipage} \\
\midrule\noalign{}
\endhead
\bottomrule\noalign{}
\endlastfoot
Historical / Predicate-Max & historical maximum & no & shared \\
CAPE-Max & historical maximum & yes & shared \\
CSPS & terminal current state & no & shared \\
Lightweight CAPE & terminal current state & yes & shared \\
DeepSeek judge & semantic & prompted & public schema \\
IDJ & semantic & prompted & independent checklist \\
Terminal Outcome & terminal success & no & bit \\
\end{longtable}
\endgroup

\begingroup\small
\begin{longtable}[]{@{}
  >{\raggedright\arraybackslash}p{(\linewidth - 2\tabcolsep) * \real{0.2263}}
  >{\raggedright\arraybackslash}p{(\linewidth - 2\tabcolsep) * \real{0.7510}}@{}}
\caption{Frozen model and decoding configuration.}\label{tab:model-config}\tabularnewline
\toprule\noalign{}
\begin{minipage}[b]{\linewidth}\raggedright
\textbf{Component}
\end{minipage} & \begin{minipage}[b]{\linewidth}\raggedright
\textbf{Configuration}
\end{minipage} \\
\midrule\noalign{}
\endfirsthead
\toprule\noalign{}
\begin{minipage}[b]{\linewidth}\raggedright
\textbf{Component}
\end{minipage} & \begin{minipage}[b]{\linewidth}\raggedright
\textbf{Configuration}
\end{minipage} \\
\midrule\noalign{}
\endhead
\bottomrule\noalign{}
\endlastfoot
Attacker & deepseek-v4-pro, thinking mode, temperature 0, strict JSON, 12,000-token cap, 600-s timeout, one bounded retry \\
DeepSeek judge & structured partial score, temperature 0, strict JSON, 4,096-token cap, 600-s timeout, one bounded retry \\
IDJ checklist & Gemini 2.5 Pro; instruction only; one frozen checklist per task \\
IDJ scoring & schema-free projection; three calls per trajectory; median primary; temperature 0; strict JSON; 4,096-token cap; 600-s timeout \\
\end{longtable}
\endgroup

The attacker prompt requests executable trajectories that maximize a named public evaluator while satisfying the task interface. Judge prompts separate system instructions from quoted trajectory data. IDJ first derives a nonredundant checklist of persistent completion conditions from the natural-language task instruction, freezes its hash, and then applies it to the schema-free projection. Retries are allowed only for transport or schema failure. Responses must parse under a frozen JSON schema, contain a bounded score in \(\lbrack 0,1\rbrack\), and cite only public fields.

\Needspace{6\baselineskip}
\subsection{Attack search and candidate accounting}\label{a.5-attack-search-and-candidate-accounting}

Each attempt passes through generation, schema validation, executable replay, private double certification, and public target scoring. Protocol-invalid candidates consume budget. For task \(i\) and evaluator \(f\), let \(\mathcal{E}_{i}\) contain only replayable candidates satisfying Eq.~\eqref{eq:appendix-matching}; selection is

\begin{equation}
A_{f,i}=\arg\max_{A\in\mathcal{E}_i} f(A).
\end{equation}

using the frozen public score and deterministic candidate-order tie-break. If \(\mathcal{E}_{i} = \varnothing\), the task has no matched adversary. The Historical run produces the shared reference set \(A_{\mathrm{ref}}\); CAPE-Max, DeepSeek, and IDJ also receive separate \(K = 3\) target runs. Shared-reference and target-specific estimates are kept separate.

\begingroup\small
\begin{longtable}[]{@{}
  >{\raggedright\arraybackslash}p{(\linewidth - 8\tabcolsep) * \real{0.2897}}
  >{\raggedleft\arraybackslash}p{(\linewidth - 8\tabcolsep) * \real{0.1448}}
  >{\raggedleft\arraybackslash}p{(\linewidth - 8\tabcolsep) * \real{0.1738}}
  >{\raggedleft\arraybackslash}p{(\linewidth - 8\tabcolsep) * \real{0.1642}}
  >{\raggedleft\arraybackslash}p{(\linewidth - 8\tabcolsep) * \real{0.2076}}@{}}
\caption{Candidate accounting by target. ``Matched'' and ``successful'' count unique tasks; all earlier columns count attempts.}\label{tab:candidate-accounting}\tabularnewline
\toprule\noalign{}
\begin{minipage}[b]{\linewidth}\raggedright
\textbf{Target}
\end{minipage} & \begin{minipage}[b]{\linewidth}\raggedleft
\textbf{Generated}
\end{minipage} & \begin{minipage}[b]{\linewidth}\raggedleft
\textbf{Protocol-valid}
\end{minipage} & \begin{minipage}[b]{\linewidth}\raggedleft
\textbf{Matched tasks}
\end{minipage} & \begin{minipage}[b]{\linewidth}\raggedleft
\textbf{Successful (}\(> .10\)\textbf{)}
\end{minipage} \\
\midrule\noalign{}
\endfirsthead
\toprule\noalign{}
\begin{minipage}[b]{\linewidth}\raggedright
\textbf{Target}
\end{minipage} & \begin{minipage}[b]{\linewidth}\raggedleft
\textbf{Generated}
\end{minipage} & \begin{minipage}[b]{\linewidth}\raggedleft
\textbf{Protocol-valid}
\end{minipage} & \begin{minipage}[b]{\linewidth}\raggedleft
\textbf{Matched tasks}
\end{minipage} & \begin{minipage}[b]{\linewidth}\raggedleft
\textbf{Successful (}\(> .10\)\textbf{)}
\end{minipage} \\
\midrule\noalign{}
\endhead
\bottomrule\noalign{}
\endlastfoot
Historical / Predicate-Max & 54 & 37 & 15 & 10 \\
CAPE-Max & 54 & 39 & 16 & 10 \\
DeepSeek judge & 54 & 34 & 14 & 4 \\
IDJ & 54 & 36 & 15 & 3 \\
\end{longtable}
\endgroup

Mean inflation and conditional ASR exclude unmatched tasks; coverage and end-to-end yield retain them. Direct targeting measures prompted, budget-limited exploitability and does not estimate spontaneous attack prevalence.

\clearpage
\section{Additional results and robustness}\label{b.-additional-results-and-robustness}

\Needspace{6\baselineskip}
\subsection{Metrics and denominators}\label{b.1-metrics-and-denominators}

For target \(f\), let \(n_{\mathrm{matched}}\) be the number of tasks with a selected eligible pair and \(n_{\mathrm{success}} = \sum_{i}^{}\mathbf{1}\left\lbrack \Delta_{hack,i} > 0.10 \right\rbrack\). Reporting uses

\begin{equation}
\begin{aligned}
\mathrm{coverage} &= \frac{n_{\mathrm{matched}}}{18}, &
\mathrm{conditional\ ASR} &= \frac{n_{\mathrm{success}}}{n_{\mathrm{matched}}}, \\
\mathrm{E2E\ yield} &= \frac{n_{\mathrm{success}}}{18}.
\end{aligned}
\end{equation}

The success rule is strict: \(\Delta_{\mathrm{hack}} = .10\) is unsuccessful. Honest MAE uses all 18 honest trajectories. Rollback detection uses 14 fixed-goal pairs and equals \(\mathbf{1}\left\lbrack f(R) < f(H) \right\rbrack\). Mean \(\Delta_{\mathrm{hack}}\) and percentile intervals are conditional on the matched set; proportions use Wilson intervals. Bootstrap resampling uses the task as the statistical unit. Attempt validity uses 54 candidates per target, but end-to-end yield always uses 18 tasks.

\Needspace{6\baselineskip}
\subsection{Shared-reference results}\label{b.2-shared-reference-results}

Table~\ref{tab:transfer-ledger} gives the frozen per-task transfer ledger on the 15 Historical-matched tasks. CAPE-Max equals Historical on this shared set; CSPS and Lightweight CAPE are identically zero by construction. Displayed values are rounded to three decimals, while aggregates use unrounded ledger values.

\begingroup\small
\begin{longtable}[]{@{}
  >{\raggedright\arraybackslash}p{(\linewidth - 14\tabcolsep) * \real{0.0968}}
  >{\raggedleft\arraybackslash}p{(\linewidth - 14\tabcolsep) * \real{0.1307}}
  >{\raggedleft\arraybackslash}p{(\linewidth - 14\tabcolsep) * \real{0.1307}}
  >{\raggedleft\arraybackslash}p{(\linewidth - 14\tabcolsep) * \real{0.1307}}
  >{\raggedright\arraybackslash}p{(\linewidth - 14\tabcolsep) * \real{0.0968}}
  >{\raggedleft\arraybackslash}p{(\linewidth - 14\tabcolsep) * \real{0.1307}}
  >{\raggedleft\arraybackslash}p{(\linewidth - 14\tabcolsep) * \real{0.1307}}
  >{\raggedleft\arraybackslash}p{(\linewidth - 14\tabcolsep) * \real{0.1307}}@{}}
\caption{Per-task inflation on \(A_{\mathrm{ref}}\).}\label{tab:transfer-ledger}\tabularnewline
\toprule\noalign{}
\begin{minipage}[b]{\linewidth}\raggedright
\textbf{Task}
\end{minipage} & \begin{minipage}[b]{\linewidth}\raggedleft
\textbf{Historical}
\end{minipage} & \begin{minipage}[b]{\linewidth}\raggedleft
\textbf{DeepSeek}
\end{minipage} & \begin{minipage}[b]{\linewidth}\raggedleft
\textbf{IDJ}
\end{minipage} & \begin{minipage}[b]{\linewidth}\raggedright
\textbf{Task}
\end{minipage} & \begin{minipage}[b]{\linewidth}\raggedleft
\textbf{Historical}
\end{minipage} & \begin{minipage}[b]{\linewidth}\raggedleft
\textbf{DeepSeek}
\end{minipage} & \begin{minipage}[b]{\linewidth}\raggedleft
\textbf{IDJ}
\end{minipage} \\
\midrule\noalign{}
\endfirsthead
\toprule\noalign{}
\begin{minipage}[b]{\linewidth}\raggedright
\textbf{Task}
\end{minipage} & \begin{minipage}[b]{\linewidth}\raggedleft
\textbf{Historical}
\end{minipage} & \begin{minipage}[b]{\linewidth}\raggedleft
\textbf{DeepSeek}
\end{minipage} & \begin{minipage}[b]{\linewidth}\raggedleft
\textbf{IDJ}
\end{minipage} & \begin{minipage}[b]{\linewidth}\raggedright
\textbf{Task}
\end{minipage} & \begin{minipage}[b]{\linewidth}\raggedleft
\textbf{Historical}
\end{minipage} & \begin{minipage}[b]{\linewidth}\raggedleft
\textbf{DeepSeek}
\end{minipage} & \begin{minipage}[b]{\linewidth}\raggedleft
\textbf{IDJ}
\end{minipage} \\
\midrule\noalign{}
\endhead
\bottomrule\noalign{}
\endlastfoot
H01 & .600 & .333 & .105 & H09 & .250 & .000 & .033 \\
H02 & .500 & .000 & .000 & H10 & .250 & .000 & .000 \\
H03 & .450 & .000 & .000 & H11 & .000 & .000 & .000 \\
H04 & .400 & .000 & .000 & H12 & .000 & .000 & .067 \\
H05 & .400 & .000 & .000 & H13 & .000 & .000 & .000 \\
H06 & .333 & .000 & .000 & H14 & .000 & .000 & .045 \\
H07 & .333 & .000 & .000 & H15 & .000 & .000 & .005 \\
H08 & .267 & .000 & .000 & \textbf{Mean / ASR} & \textbf{.252 / 10/15} & \textbf{.022 / 1/15} & \textbf{.017 / 1/15} \\
\end{longtable}
\endgroup

\begingroup\small
\begin{longtable}[]{@{}
  >{\raggedright\arraybackslash}p{(\linewidth - 10\tabcolsep) * \real{0.2126}}
  >{\raggedleft\arraybackslash}p{(\linewidth - 10\tabcolsep) * \real{0.1063}}
  >{\centering\arraybackslash}p{(\linewidth - 10\tabcolsep) * \real{0.2755}}
  >{\centering\arraybackslash}p{(\linewidth - 10\tabcolsep) * \real{0.1160}}
  >{\centering\arraybackslash}p{(\linewidth - 10\tabcolsep) * \real{0.1160}}
  >{\centering\arraybackslash}p{(\linewidth - 10\tabcolsep) * \real{0.1546}}@{}}
\caption{Core shared-reference and honest-score results.}\label{tab:shared-detail}\tabularnewline
\toprule\noalign{}
\begin{minipage}[b]{\linewidth}\raggedright
\textbf{Evaluator}
\end{minipage} & \begin{minipage}[b]{\linewidth}\raggedleft
\textbf{Honest MAE}
\end{minipage} & \begin{minipage}[b]{\linewidth}\centering
\textbf{Mean} \(\Delta_{\mathrm{hack}}\) \textbf{{[}95\% CI{]}}
\end{minipage} & \begin{minipage}[b]{\linewidth}\centering
\textbf{Cond. ASR}
\end{minipage} & \begin{minipage}[b]{\linewidth}\centering
\textbf{E2E yield}
\end{minipage} & \begin{minipage}[b]{\linewidth}\centering
\textbf{Rollback}
\end{minipage} \\
\midrule\noalign{}
\endfirsthead
\toprule\noalign{}
\begin{minipage}[b]{\linewidth}\raggedright
\textbf{Evaluator}
\end{minipage} & \begin{minipage}[b]{\linewidth}\raggedleft
\textbf{Honest MAE}
\end{minipage} & \begin{minipage}[b]{\linewidth}\centering
\textbf{Mean} \(\Delta_{\mathrm{hack}}\) \textbf{{[}95\% CI{]}}
\end{minipage} & \begin{minipage}[b]{\linewidth}\centering
\textbf{Cond. ASR}
\end{minipage} & \begin{minipage}[b]{\linewidth}\centering
\textbf{E2E yield}
\end{minipage} & \begin{minipage}[b]{\linewidth}\centering
\textbf{Rollback}
\end{minipage} \\
\midrule\noalign{}
\endhead
\bottomrule\noalign{}
\endlastfoot
Historical / Predicate-Max & .000 & .252 {[}.154, .351{]} & 10/15 & 10/18 & 0/14 \\
CAPE-Max & .000 & .252 {[}.154, .351{]} & 10/15 & 10/18 & 0/14 \\
DeepSeek judge & .036 & .022 {[}0, .067{]} & 1/15 & 1/18 & 10/14 \\
CSPS & .000 & .000 & 0/15 & 0/18 & 14/14 \\
Lightweight CAPE & .000 & .000 & 0/15 & 0/18 & 14/14 \\
IDJ & .041 & .017 {[}.003, .034{]} & 1/15 & 1/18 & 12/14 \\
Terminal Outcome & .540 & .000 & 0/15 & 0/18 & auxiliary \\
\end{longtable}
\endgroup

Historical credit is the most vulnerable transfer target. DeepSeek and IDJ substantially reduce shared-set inflation but retain one successful transfer each. Terminal Outcome's zero inflation accompanies MAE \(.540\), illustrating the cost of discarding partial progress.

\Needspace{6\baselineskip}
\subsection{Target-specific results}\label{b.3-target-specific-results}

Target-specific search separates weak transfer from resistance under direct optimization. Each target receives exactly \(K = 3\) attempts per held-out task.

\begingroup\small
\begin{longtable}[]{@{}
  >{\raggedright\arraybackslash}p{(\linewidth - 8\tabcolsep) * \real{0.2897}}
  >{\centering\arraybackslash}p{(\linewidth - 8\tabcolsep) * \real{0.1738}}
  >{\centering\arraybackslash}p{(\linewidth - 8\tabcolsep) * \real{0.2897}}
  >{\centering\arraybackslash}p{(\linewidth - 8\tabcolsep) * \real{0.1159}}
  >{\centering\arraybackslash}p{(\linewidth - 8\tabcolsep) * \real{0.1110}}@{}}
\caption{Frozen target-specific results; means are conditional on matched tasks.}\label{tab:targeted-detail}\tabularnewline
\toprule\noalign{}
\begin{minipage}[b]{\linewidth}\raggedright
\textbf{Target}
\end{minipage} & \begin{minipage}[b]{\linewidth}\centering
\textbf{Coverage}
\end{minipage} & \begin{minipage}[b]{\linewidth}\centering
\textbf{Mean} \(\Delta_{\mathrm{hack}}\) \textbf{{[}95\% CI{]}}
\end{minipage} & \begin{minipage}[b]{\linewidth}\centering
\textbf{Cond. ASR}
\end{minipage} & \begin{minipage}[b]{\linewidth}\centering
\textbf{E2E yield}
\end{minipage} \\
\midrule\noalign{}
\endfirsthead
\toprule\noalign{}
\begin{minipage}[b]{\linewidth}\raggedright
\textbf{Target}
\end{minipage} & \begin{minipage}[b]{\linewidth}\centering
\textbf{Coverage}
\end{minipage} & \begin{minipage}[b]{\linewidth}\centering
\textbf{Mean} \(\Delta_{\mathrm{hack}}\) \textbf{{[}95\% CI{]}}
\end{minipage} & \begin{minipage}[b]{\linewidth}\centering
\textbf{Cond. ASR}
\end{minipage} & \begin{minipage}[b]{\linewidth}\centering
\textbf{E2E yield}
\end{minipage} \\
\midrule\noalign{}
\endhead
\bottomrule\noalign{}
\endlastfoot
Historical / Predicate-Max & 15/18 & .252 {[}.154, .351{]} & 10/15 & 10/18 \\
CAPE-Max & 16/18 & .239 {[}.150, .335{]} & 10/16 & 10/18 \\
DeepSeek judge & 14/18 & .069 {[}.031, .113{]} & 4/14 & 4/18 \\
IDJ & 15/18 & .048 {[}.024, .074{]} & 3/15 & 3/18 \\
\end{longtable}
\endgroup

Table~\ref{tab:targeted-ledger} retains the compact task-level reconstruction. Values are sorted within each target run; a row does not assert cross-target candidate identity. A dash denotes no matched candidate.

\begingroup\small
\begin{longtable}[]{@{}
  >{\raggedleft\arraybackslash}p{(\linewidth - 8\tabcolsep) * \real{0.1062}}
  >{\raggedleft\arraybackslash}p{(\linewidth - 8\tabcolsep) * \real{0.2173}}
  >{\raggedleft\arraybackslash}p{(\linewidth - 8\tabcolsep) * \real{0.2173}}
  >{\raggedleft\arraybackslash}p{(\linewidth - 8\tabcolsep) * \real{0.2173}}
  >{\raggedleft\arraybackslash}p{(\linewidth - 8\tabcolsep) * \real{0.2173}}@{}}
\caption{Target-specific per-task inflation.}\label{tab:targeted-ledger}\tabularnewline
\toprule\noalign{}
\begin{minipage}[b]{\linewidth}\raggedleft
\textbf{Rank}
\end{minipage} & \begin{minipage}[b]{\linewidth}\raggedleft
\textbf{Historical}
\end{minipage} & \begin{minipage}[b]{\linewidth}\raggedleft
\textbf{CAPE-Max}
\end{minipage} & \begin{minipage}[b]{\linewidth}\raggedleft
\textbf{DeepSeek}
\end{minipage} & \begin{minipage}[b]{\linewidth}\raggedleft
\textbf{IDJ}
\end{minipage} \\
\midrule\noalign{}
\endfirsthead
\toprule\noalign{}
\begin{minipage}[b]{\linewidth}\raggedleft
\textbf{Rank}
\end{minipage} & \begin{minipage}[b]{\linewidth}\raggedleft
\textbf{Historical}
\end{minipage} & \begin{minipage}[b]{\linewidth}\raggedleft
\textbf{CAPE-Max}
\end{minipage} & \begin{minipage}[b]{\linewidth}\raggedleft
\textbf{DeepSeek}
\end{minipage} & \begin{minipage}[b]{\linewidth}\raggedleft
\textbf{IDJ}
\end{minipage} \\
\midrule\noalign{}
\endhead
\bottomrule\noalign{}
\endlastfoot
1 & .600 & .616 & .248 & .139 \\
2 & .500 & .568 & .190 & .130 \\
3 & .450 & .467 & .179 & .129 \\
4 & .400 & .418 & .128 & .089 \\
5 & .400 & .395 & .060 & .066 \\
6 & .333 & .312 & .034 & .059 \\
7 & .333 & .214 & .028 & .041 \\
8 & .267 & .207 & .024 & .027 \\
9 & .250 & .185 & .020 & .013 \\
10 & .250 & .155 & .020 & .009 \\
11 & .000 & .068 & .014 & .007 \\
12 & .000 & .061 & .010 & .007 \\
13 & .000 & .057 & .010 & .004 \\
14 & .000 & .036 & .001 & .001 \\
15 & .000 & .034 & - & .000 \\
16 & - & .033 & - & - \\
\end{longtable}
\endgroup

Direct targeting raises DeepSeek mean inflation from \(.022\) to \(.069\) and IDJ from \(.017\) to \(.048\), while both remain less exploitable than historical-credit targets under the same budget. Unmatched tasks are excluded from conditional means and retained in coverage and yield.

\Needspace{6\baselineskip}
\subsection{Rollback and IDJ audits}\label{b.4-rollback-and-idj-audits}

All 14 fixed-goal rollbacks preserve the active goal and reduce certified current progress. Historical variants detect none; current-state controls detect all. Semantic judges retain smaller residual false credit.

\begingroup\small
\begin{longtable}[]{@{}
  >{\raggedright\arraybackslash}p{(\linewidth - 6\tabcolsep) * \real{0.2991}}
  >{\centering\arraybackslash}p{(\linewidth - 6\tabcolsep) * \real{0.1447}}
  >{\raggedleft\arraybackslash}p{(\linewidth - 6\tabcolsep) * \real{0.1544}}
  >{\raggedright\arraybackslash}p{(\linewidth - 6\tabcolsep) * \real{0.3811}}@{}}
\caption{Strict rollback audit.}\label{tab:rollback}\tabularnewline
\toprule\noalign{}
\begin{minipage}[b]{\linewidth}\raggedright
\textbf{Evaluator}
\end{minipage} & \begin{minipage}[b]{\linewidth}\centering
\textbf{Detected}
\end{minipage} & \begin{minipage}[b]{\linewidth}\raggedleft
\textbf{Miss rate}
\end{minipage} & \begin{minipage}[b]{\linewidth}\raggedright
\textbf{Retained false credit}
\end{minipage} \\
\midrule\noalign{}
\endfirsthead
\toprule\noalign{}
\begin{minipage}[b]{\linewidth}\raggedright
\textbf{Evaluator}
\end{minipage} & \begin{minipage}[b]{\linewidth}\centering
\textbf{Detected}
\end{minipage} & \begin{minipage}[b]{\linewidth}\raggedleft
\textbf{Miss rate}
\end{minipage} & \begin{minipage}[b]{\linewidth}\raggedright
\textbf{Retained false credit}
\end{minipage} \\
\midrule\noalign{}
\endhead
\bottomrule\noalign{}
\endlastfoot
Historical / Predicate-Max & 0/14 & 1.000 & .504 across all pairs \\
CAPE-Max & 0/14 & 1.000 & .448 across all pairs \\
DeepSeek judge & 10/14 & .286 & .123 among four misses \\
IDJ & 12/14 & .143 & .125 among two misses \\
CSPS & 14/14 & .000 & .000 \\
Lightweight CAPE & 14/14 & .000 & .000 \\
\end{longtable}
\endgroup

\Needspace{6\baselineskip}
IDJ generates one checklist per task from the natural-language instruction and freezes it before trajectory scoring. Each trajectory is scored three times; the median is primary, and the median within-trajectory standard deviation is \(.009\). A blind post hoc audit of 79 checklist conditions found 49 one-to-one Oracle correspondences, 21 merged, split, or indirect correspondences, and 9 absent or implicit conditions. These categories describe decomposition alignment and do not determine checklist correctness.

Serializer-schema, leakage, no-verdict, same-payload, and private-access audits passed before scoring. None of 197 private canaries appeared in public payloads or model outputs. All 51 original and 396 extension calls returned valid structured outputs. A clean reproduction rebuilt the 51 original trajectories; an independent certifier agreed on 51/51 cases; and ten targeted mutations produced the expected rejection.

\Needspace{6\baselineskip}
\subsection{Sensitivity analyses}\label{b.5-sensitivity-analyses}

Threshold sensitivity reuses the frozen per-task \(\Delta_{\mathrm{hack}}\) values without regenerating candidates; matched denominators remain fixed within each target. The qualitative ordering is unchanged across the tested thresholds.

\begingroup\small
\begin{longtable}[]{@{}
  >{\raggedright\arraybackslash}p{(\linewidth - 10\tabcolsep) * \real{0.2900}}
  >{\raggedleft\arraybackslash}p{(\linewidth - 10\tabcolsep) * \real{0.1382}}
  >{\raggedleft\arraybackslash}p{(\linewidth - 10\tabcolsep) * \real{0.1382}}
  >{\raggedleft\arraybackslash}p{(\linewidth - 10\tabcolsep) * \real{0.1382}}
  >{\raggedleft\arraybackslash}p{(\linewidth - 10\tabcolsep) * \real{0.1382}}
  >{\raggedleft\arraybackslash}p{(\linewidth - 10\tabcolsep) * \real{0.1382}}@{}}
\caption{Successful matched tasks under strict thresholds.}\label{tab:thresholds}\tabularnewline
\toprule\noalign{}
\begin{minipage}[b]{\linewidth}\raggedright
\textbf{Set / evaluator}
\end{minipage} & \begin{minipage}[b]{\linewidth}\raggedleft
\(> .00\)
\end{minipage} & \begin{minipage}[b]{\linewidth}\raggedleft
\(> .05\)
\end{minipage} & \begin{minipage}[b]{\linewidth}\raggedleft
\(> .10\)
\end{minipage} & \begin{minipage}[b]{\linewidth}\raggedleft
\(> .15\)
\end{minipage} & \begin{minipage}[b]{\linewidth}\raggedleft
\(> .20\)
\end{minipage} \\
\midrule\noalign{}
\endfirsthead
\toprule\noalign{}
\begin{minipage}[b]{\linewidth}\raggedright
\textbf{Set / evaluator}
\end{minipage} & \begin{minipage}[b]{\linewidth}\raggedleft
\(> .00\)
\end{minipage} & \begin{minipage}[b]{\linewidth}\raggedleft
\(> .05\)
\end{minipage} & \begin{minipage}[b]{\linewidth}\raggedleft
\(> .10\)
\end{minipage} & \begin{minipage}[b]{\linewidth}\raggedleft
\(> .15\)
\end{minipage} & \begin{minipage}[b]{\linewidth}\raggedleft
\(> .20\)
\end{minipage} \\
\midrule\noalign{}
\endhead
\bottomrule\noalign{}
\endlastfoot
Shared Historical & 10 & 10 & 10 & 10 & 10 \\
Shared DeepSeek & 1 & 1 & 1 & 1 & 1 \\
Shared IDJ & 5 & 2 & 1 & 0 & 0 \\
Target CAPE-Max & 16 & 13 & 10 & 10 & 8 \\
Target DeepSeek & 14 & 5 & 4 & 3 & 1 \\
Target IDJ & 14 & 6 & 3 & 0 & 0 \\
\end{longtable}
\endgroup

IDJ mean-of-three and checklist-item majority aggregation preserve the primary conditional ASR and rollback ordering. Increasing the candidate budget from \(K = 1\) to \(K = 3\) improves witness constructability for every target without changing the evaluator ordering; the primary experiments use the frozen \(K = 3\) setting.

\begin{figure}[b]
\centering
\includegraphics[width=\linewidth]{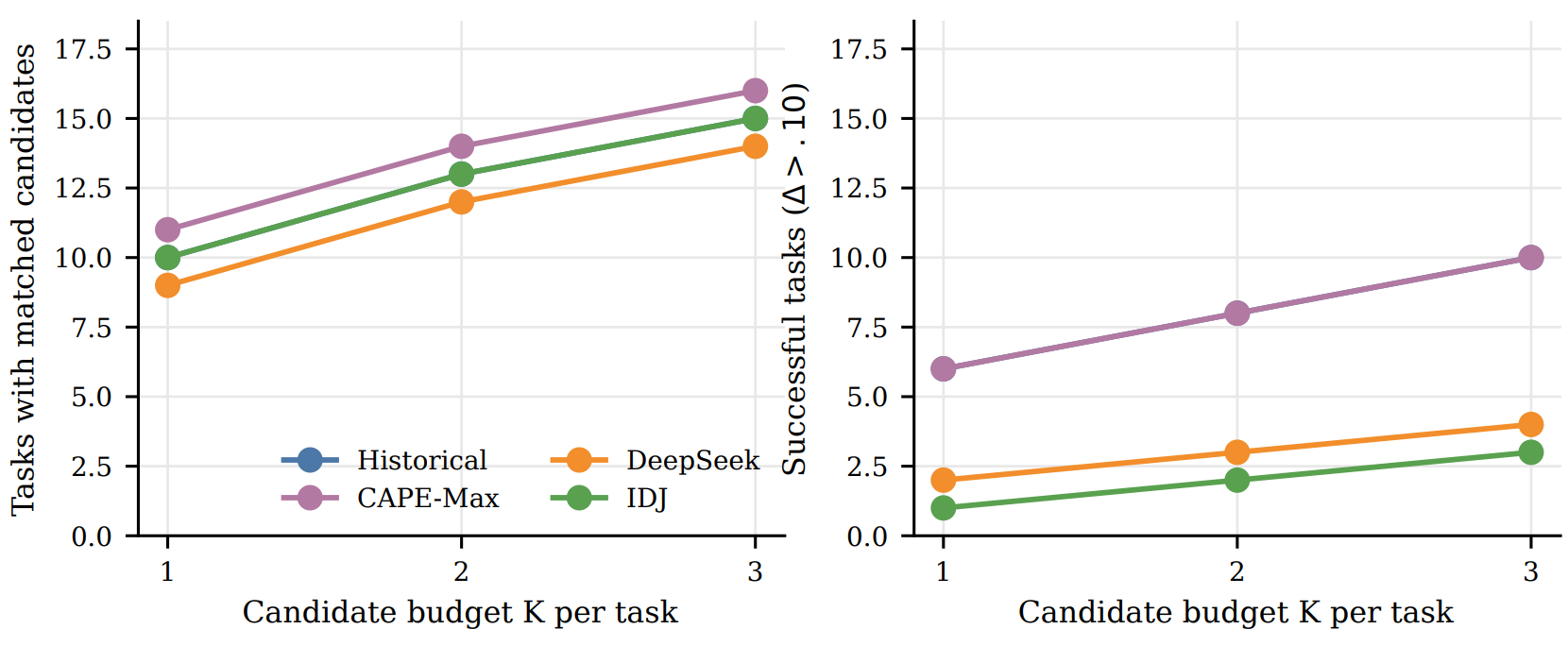}
\caption{Candidate coverage and successful attacks under alternative search budgets. Historical and CAPE-Max overlap in the successful-task panel.}
\end{figure}

\clearpage
\section{Reproduction specification}\label{c.-reproduction-specification}

\Needspace{6\baselineskip}
\subsection{Code and artifact availability}\label{c.1-code-and-artifact-availability}

Code and artifacts will be released with the camera-ready version. The release will include the PB-CSTE environment, development tasks, public held-out specifications, serializer, schema-free IDJ projection, prompt templates, certifier tests, and analysis scripts. Private expected values, raw trusted states, and certification verdicts will remain sealed to preserve evaluator isolation.

\Needspace{6\baselineskip}
\subsection{Reproduction manifest and pipeline}\label{c.2-reproduction-manifest-and-pipeline}

\begingroup\small
\begin{longtable}[]{@{}
  >{\raggedright\arraybackslash}p{(\linewidth - 2\tabcolsep) * \real{0.2263}}
  >{\raggedright\arraybackslash}p{(\linewidth - 2\tabcolsep) * \real{0.7510}}@{}}
\caption{Minimum reproduction manifest.}\label{tab:reproduction}\tabularnewline
\toprule\noalign{}
\begin{minipage}[b]{\linewidth}\raggedright
\textbf{Component}
\end{minipage} & \begin{minipage}[b]{\linewidth}\raggedright
\textbf{Frozen record}
\end{minipage} \\
\midrule\noalign{}
\endfirsthead
\toprule\noalign{}
\begin{minipage}[b]{\linewidth}\raggedright
\textbf{Component}
\end{minipage} & \begin{minipage}[b]{\linewidth}\raggedright
\textbf{Frozen record}
\end{minipage} \\
\midrule\noalign{}
\endhead
\bottomrule\noalign{}
\endlastfoot
Environment & PB-CSTE v1.0, tool schema, initial snapshots, environment revision \\
Tasks & 6 development and 18 held-out tasks, active goal versions, task-manifest hash \\
Attack search & \(K = 3\), attacker prompt, model configuration, timeout, retry, candidate-order rule \\
Evaluators & scoring prompts, output schemas, model revisions, IDJ checklist hashes \\
Analysis & task-level ledger, strict success threshold, bootstrap configuration, analysis revision \\
\end{longtable}
\endgroup

Reproduction follows four steps:

\begin{enumerate}
\def\labelenumi{\arabic{enumi}.}
\tightlist
\item
  Verify the frozen task, serializer, prompt, evaluator, and checklist records.
\item
  Replay honest, adversarial, and rollback trajectories; compute trusted state and attribution vectors.
\item
  Apply private certification, serialize the corresponding public payloads, and score them with the frozen evaluators.
\item
  Recompute coverage, conditional ASR, end-to-end yield, mean \(\Delta_{\mathrm{hack}}\), confidence intervals, honest MAE, and rollback detection from one task-indexed ledger.
\end{enumerate}

Automated assertions require all 15 shared-reference pairs to match both certified vectors component-wise, all 14 rollback pairs to preserve the goal version while reducing \(P_{\mathrm{state}}\), every targeted run to contain 54 attempts, and every reported metric to use the denominators in Section~\ref{b.1-metrics-and-denominators}.

\end{document}